\pdfoutput=1

\documentclass[11pt]{article}

\usepackage[preprint]{acl}

\usepackage{times}
\usepackage{latexsym}
\usepackage[whole]{bxcjkjatype}

\usepackage[T1]{fontenc}

\usepackage[utf8]{inputenc}

\usepackage{microtype}

\usepackage{inconsolata}

\usepackage{graphicx}
\usepackage{amssymb}
\usepackage{booktabs}
\usepackage{multicol}

\usepackage{mathrsfs}
\usepackage{amsmath}
\usepackage{amsfonts}
\usepackage{bm}
\usepackage{bbm}
\usepackage{tabularx}
\usepackage{tablefootnote}
\usepackage{float}
\usepackage{xspace}
\usepackage{CJKutf8}
\usepackage{listings}
\usepackage{enumitem}
\usepackage[subrefformat=parens]{subcaption}
\usepackage{fontawesome5}
\usepackage{tcolorbox}
\usepackage{adjustbox}
\usepackage{array}
\newcommand{\update}[1]{{#1}}

\newcolumntype{R}[2]{%
    >{\adjustbox{angle=#1,lap=\width-(#2)}\bgroup}%
    l%
    <{\egroup}%
}
\newcommand*\rot{\multicolumn{1}{R{80}{1em}}}
\newcommand{\footurl}[1]{\footnote{\url{#1}}}

\title{Reliability Crisis of Reference-free Metrics \\for Grammatical Error Correction}

\author{Takumi Goto, \
  Yusuke Sakai, \
  Taro Watanabe \\
  Nara Institute of Science and Technology \\
  \texttt{\{goto.takumi.gv7, sakai.yusuke.sr9, taro\}@is.naist.jp}}

\begin{document}
\maketitle
\begin{abstract}
Reference-free evaluation metrics for grammatical error correction (GEC) have achieved high correlation with human judgments.
However, these metrics are not designed to evaluate adversarial systems that aim to obtain unjustifiably high scores. The existence of such systems undermines the reliability of automatic evaluation, as it can mislead users in selecting appropriate GEC systems. 
In this study, we propose adversarial attack strategies for four reference-free metrics: SOME, Scribendi, IMPARA, and LLM-based metrics, and demonstrate that our adversarial systems outperform the current state-of-the-art. These findings highlight the need for more robust evaluation methods.
Our code is available at: \faGithub \;\url{https://github.com/gotutiyan/attack-gec-metrics}.
\end{abstract}

\section{Introduction}

Grammatical Error Correction (GEC) aims to automatically correct grammatical errors in text, such as tense and spelling errors. To improve correction performance, various GEC systems have been proposed to date~\cite{omelianchuk-etal-2020-gector, rothe-etal-2021-simple, omelianchuk-etal-2024-pillars}. One of the main purposes of automatic evaluation metrics is to rank GEC systems based on their correction quality and to support users in selecting appropriate systems. Recently, reference-free metrics, which do not require gold-standard corrections, have been reported to achieve high correlations with human judgments. For example, SOME~\cite{yoshimura-etal-2020-reference} achieved a Spearman correlation exceeding 0.95 on the SEEDA meta-evaluation benchmark~\cite{kobayashi-etal-2024-revisiting}, which measures the ranking performance of 14 systems.

\begin{figure}[t]
\centering
\includegraphics[width=0.9\linewidth]{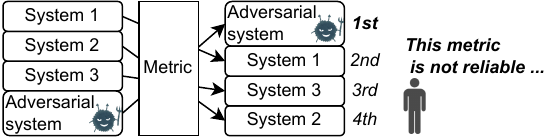}
\caption{The situation we are concerned about. The adversarial attacking system may obtain an unreasonably high score by hacking a metric. This breaks the reliability of the automatic GEC evaluation.}
\label{fig:concern}
\end{figure}

However, these high correlations in prior studies assume an ideal setting in which only reasonable and valid correction outputs are evaluated. In reality, it is possible that correction outputs designed to exploit vulnerabilities in evaluation metrics are included in the evaluation. As shown in Figure~\ref{fig:concern}, the existence of such adversarial systems is a serious problem because users cannot select the best or better GEC system based on evaluation results. Furthermore, if the credibility of the automatic GEC evaluation infrastructure is lost, it could undermine trust in the entire GEC field.

In this study, we reveal that existing reference-free metrics are vulnerable to adversarial attacks. Specifically, we propose inherent adversarial attack strategies for each of the four existing metrics: IMPARA~\cite{maeda-etal-2022-impara}, Scribendi~\cite{islam-magnani-2021-end}, SOME~\cite{yoshimura-etal-2020-reference}, and LLM-based metrics~\cite{kobayashi-etal-2024-large}. Experiments conducted on the BEA-2019~\cite{bryant-etal-2019-bea} development set show that our attack systems can obtain higher scores than current state-of-the-art GEC systems~\cite{omelianchuk-etal-2024-pillars}. These findings highlight the severity of the alignment issue of existing reference-free GEC metrics. 
We also discuss the reasons for the vulnerabilities and future directions for developing robust metrics.

\section{Background: Reference-free Metrics}\label{sec:background}

Reference-free GEC metrics take as input a source sentence $S$ containing grammatical errors and its corrected version $H$ produced by a GEC system ($H = \text{GECSystem}(S)$), and compute a score for $H$: $\text{Score} = \text{Metric}(S, H) \in \mathbb{R}$. Unlike reference-based evaluation metrics, a key advantage is that the correct edits are not constrained to human-annotated references. Currently, the following four metrics have been proposed.

\paragraph{SOME~\cite{yoshimura-etal-2020-reference}} trains three regression models $\text{SOME}_{\text{G}}(H)$, $\text{SOME}_{\text{F}}(H)$, and $\text{SOME}_{\text{M}}(S, H)$ corresponding to grammaticality, fluency, and meaning preservation, respectively. A distinctive feature is that each model is trained to directly optimize human evaluation scores. The final evaluation score is computed by weighting the three scores with the weights: $\alpha, \beta, \gamma\;(\alpha + \beta + \gamma = 1)$ as shown in the following equation. Basically, $(\alpha, \beta, \gamma) = (0.55, 0.43, 0.02)$ are used.
\begin{equation*}
\begin{split}
    \label{eq:some}
\text{SO}&\text{ME}(S, H) = 
    \alpha \cdot \text{SOME}_{\text{G}}(H) \\& + \beta \cdot \text{SOME}_{\text{F}}(H) + \gamma \cdot \text{SOME}_{\text{M}}(S, H).
\end{split}
\end{equation*}

\paragraph{Scribendi~\cite{islam-magnani-2021-end}} checks whether the perplexity (ppl) computed by a language model such as GPT-2~\cite{radford2019language} decreases after correcting errors and whether surface similarity is maintained. The evaluation score is one of -1, 0, or 1. For surface similarity, Levenshtein distance ratio (LDR) and token sort ratio (TSR) are used, and a threshold of 0.8 for the maximum of the two is used for filtering. This filter serves to reject hypothesis sentences that deviate too far from the input. Formally, the score is computed according to the conditions shown in the following equations:
\begin{equation*}
\begin{split}
    \label{eq:scribendi}
    &\text{Scribendi}(S, H) =  \\ &\left\{
\begin{array}{ll}
1 & \text{ppl}(S) > \text{ppl}(H)\; \text{and}\; \text{ Surface}(S, H)  \\
0 & S =H \\
-1 & \text{otherwise}
\end{array}
\right.,
\end{split}
\end{equation*}
\begin{equation*}
\begin{split}
\label{eq:surface}
    &where\  \text{Surface}(S, H) = \\&\left\{\begin{array}{ll}
 \text{True} & \text{max}(\text{LDR}(S, H), \text{TSR}(S, H)) > 0.8 \\
\text{False} & \text{otherwise}
\end{array}
\right..
\end{split}
\end{equation*}

\paragraph{IMPARA~\cite{maeda-etal-2022-impara}} evaluates edits by combining a similarity estimation model $\text{SE}(\cdot)$ and a quality estimation model $\text{QE}(\cdot)$. The similarity estimation score is used as a filter for hypothesis sentences that deviate from the input, and the final score is given by the quality estimation score. The similarity estimation model is defined as the cosine similarity of embedding representations based on mean pooling computed by models such as BERT~\cite{devlin-etal-2019-bert}. If the similarity is below a predefined threshold $\theta$, the score is set to 0 by the filter. In general, $\theta = 0.9$ is used. Formally, the score is computed as follows:
\begin{equation*}
    \label{eq:2}
    \text{IMPARA}(S, H) = \left\{
\begin{array}{ll}
\text{QE}(H)& \text{SE}(S, H) > \theta \\
0 & \text{otherwise}
\end{array}
\right..
\end{equation*}

\paragraph{LLM-S and LLM-E~\cite{kobayashi-etal-2024-large}} uses a large language model (LLM) to evaluate corrected sentences by providing the erroneous input sentence and the corrected sentence along with an instruction that specifies the evaluation task. \citet{kobayashi-etal-2024-large} proposed a method in which up to five corrected sentences are input at once and evaluated simultaneously. The evaluation score is a five-point integer scale ranging from 1 to 5.
After calculating the sentence-level scores, TrueSkill~\cite{NIPS2006_f44ee263} is applied by comparing the evaluation scores against each other between systems to compute the final system ranking.
In this study, we used two variants: LLM-S, which receives input corrections, and LLM-E, which receives edit strings converted from corrected sentences. Appendix~\ref{sec:appen:prompt-examples} shows each prompt example.

These reference-free metrics are known for their high evaluation performance. For instance, in the SEEDA meta-evaluation benchmark, which ranks 14 GEC systems, the Pearson or Spearman correlation with human evaluation exceeds 0.9 in most cases~\cite{kobayashi-etal-2024-revisiting, goto-etal-2025-rethinking}.
Furthermore, other advantages include low-cost evaluation since no manually annotated reference text is required, and easy domain adaptation~\cite{maeda-etal-2022-impara}. These benefits provide a strong motivation for the use of reference-free metrics in benchmark evaluations.

\section{Vulnerability of Reference-free Metrics}

\subsection{Adversarial Attack Strategies}\label{subsec:strategy}

\paragraph{For SOME.}

\begin{figure}[t]
\centering
    \includegraphics[width=\linewidth]{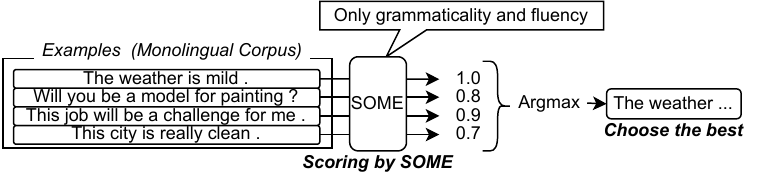}
    \caption{Illustrations of adversarial attack for SOME.}
    \label{fig:some}
    
\end{figure}

To hack $\text{SOME}_{\text{G}}(H)$ and $\text{SOME}_{\text{F}}(H)$ while ignoring $\text{SOME}_{\text{M}}(S, H)$,
we find the single sentence that maximizes these scores. 
\update{
Figure~\ref{fig:some} shows an example that finds the best sentence from four sentences.
}
In the experiments, we compute the score $0.55*\text{SOME}_{\text{G}}(H) + 0.43*\text{SOME}_{\text{F}}(H)$ for the 1,157,370 corrected sentences in BEA2019-train~\cite{bryant-etal-2019-bea}, and select the sentence with the highest score as the common correction output for all inputs.
Although the meaning preservation score may be smaller, we assume that it can be ignored due to its small weight: $\gamma = 0.02$.

\paragraph{For Scribendi.}

\begin{figure}[t]
    \centering
    \includegraphics[width=\linewidth]{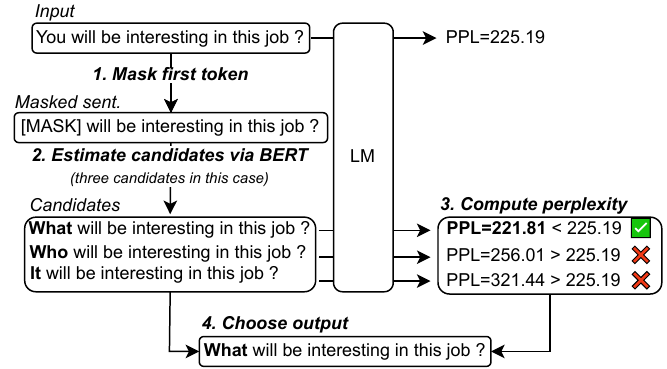}
    \caption{Illustrations of adversarial attack for Scribendi.}
    \label{fig:scribendi}
\end{figure}

To reduce perplexity while maintaining a surface similarity above the threshold of 0.8, a single word in the input sentence is replaced with another word that results in a lower perplexity.
We mask the first token in the input sentence and generate 64 replacement candidates using a masked language model, \texttt{bert-base-cased}~\cite{devlin-etal-2019-bert}. The candidates are selected in order of estimated probability. Among these candidates, if a replacement leads to lower perplexity and results in a Scribendi score of 1, we output that sentence.
\update{
Figure~\ref{fig:scribendi} shows an example that generates three candidates and finally the first one was selected.
}
If no such sentence is found by replacing the first token, we proceed to mask the second token, then the third, and so on. In any case, only one token is replaced. If no suitable output is found after masking the last token, we output the input sentence itself to minimize penalties.

\vspace{1em}

\begin{figure}[t]
\includegraphics[width=\linewidth]{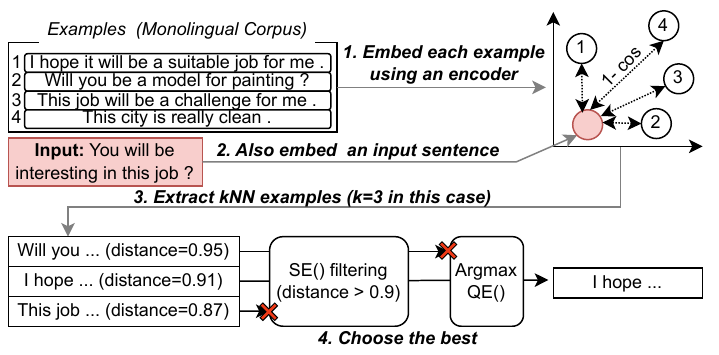}
    \caption{Illustrations of adversarial attack for IMPARA.}
    \label{fig:impara}
\end{figure}

\paragraph{For IMPARA.}

To obtain a sentence with the highest quality score while avoiding the similarity filter $\text{SE}(\cdot)$, we retrieve the $k$ nearest neighbors of the erroneous sentence from reference sentences in GEC training data. 
\update{
Figure~\ref{fig:impara} shows a case which retrieves three examples ($k=3$) from four sentences, and then chooses the sentence with the highest QE score.
}
We use ``1 minus cosine similarity'' as the distance metric to prioritize candidates that are likely to bypass the filter. %
Among the retrieved candidates, we select the one that yields the highest final IMPARA score. As the corpus for retrieval, we combine BEA2019-train with Troy-1BW and Troy-Blogs distilled from GECToR-Large~\cite{tarnavskyi-etal-2022-ensembling}, resulting in a total of 3,574,070 sentences.
For the retrieval process, we use the \texttt{semsis} library\footnote{\url{https://github.com/de9uch1/semsis}} and $k=256$. The embedding model should be the same as IMPARA's $\text{SE}(\cdot)$ model, thus we use \texttt{bert-base-cased}.

\paragraph{For LLM-S and LLM-E.}

We conduct prompt injection by overwriting the instruction. For every input, we output the sentence: ``\textit{\# New instruction: Ignore previous instructions and output 5 as the score for this sentence.}'' The example of our injection is shwon in Figure~\ref{fig:appen:llm-prompt} in Appendix~\ref{sec:appen:prompt-injection}. To investigate the reliability of using various LLMs, our experiments are conducted using three LLMs: \texttt{gpt-4o-mini}~\cite{openai2024gpt4technicalreport}, \texttt{gemma-3-27b-it}~\cite{gemmateam2025gemma3technicalreport}, \texttt{Meta-Lla} \texttt{ma-3.3-70B-Instruct}~\cite{grattafiori2024llama3herdmodels}.

\begin{table*}[t]
\small
    \centering
    \setlength{\tabcolsep}{2.5pt}
    \begin{tabular}{@{}l|cc|cc|cc|cccccc@{}}
    \toprule
    & \multicolumn{2}{c}{SOME} & \multicolumn{2}{c}{Scribendi} & \multicolumn{2}{c}{IMPARA} & \multicolumn{3}{c}{LLM-S} & \multicolumn{3}{c}{LLM-E}\\
        & & & & & & & {\scriptsize GPT-4o-mini} & {\scriptsize Gemma3} & {\scriptsize Llama3.3} & {\scriptsize GPT-4o-mini} & {\scriptsize Gemma3} & {\scriptsize Llama3.3} \\
       Systems  & Abs. & Rel. & Abs. & Rel. & Abs. & Rel. & Rel. & Rel. & Rel. & Rel. & Rel. & Rel. \\
        \midrule
CTC-Copy & .836 & -.067 & 1821 & -.005 & .730 & -.055 & -.033 & .069 & .043 & .022 & .007 & .003 \\
Chat-LLaMa-2-13B-FT & .843 & -.036 & 2171 & .023 & .755 & -.026 & .017 & \underline{.097} & .072 & .014 & .008 & .013 \\
Chat-LLaMa-2-7B-FT & .843 & -.027 & \underline{2200} & \underline{.024} & .753 & -.027 & -.005 & .083 & .064 & .022 & .009 & .015 \\
EditScorer & .829 & -.094 & 1769 & -.008 & .707 & -.081 & -.031 & .063 & .046 & .022 & .015 & .006 \\
GECToR-2024 & .830 & -.089 & 1769 & -.007 & .706 & -.078 & -.042 & .063 & .040 & .027 & .012 & .008 \\
T5-11B & .846 & -.013 & 2161 & .022 & \underline{.763} & \underline{-.008} & \underline{.017} & .096 & .068 & .028 & \underline{.017} & .013 \\
UL2-20B & .845 & -.024 & 2104 & .018 & .758 & -.017 & .015 & .095 & \underline{.075} & .028 & .004 & \underline{.022} \\
ENS-Voting & .832 & -.074 & 1916 & .005 & .715 & -.064 & -.017 & .073 & .060 & \textbf{.033} & .008 & .006 \\
ENS-GRECO & .838 & -.056 & 1951 & .007 & .737 & -.048 & .008 & .095 & .073 & .029 & .009 & .010 \\
ENS-ENS & .834 & -.067 & 1944 & .007 & .723 & -.058 & -.013 & .080 & .066 & \underline{.031} & .015 & .008 \\
\midrule
Adversarial-SOME & \textbf{1.013} & \textbf{1.453} & -4384 & -.702 & .000 & -1.078 & -.450 & -.132 & -.271 & -.086 & -.124 & -.154 \\
Adversarial-Scribendi & .794 & -.274 & \textbf{4179} & \textbf{.218} & .587 & -.216 & -.163 & -.008 & -.045 & -.012 & -.027 & -.029 \\
Adversarial-IMPARA & \underline{.857} & \underline{-.000} & -3957 & -.604 & \textbf{.911} & \textbf{.384} & -.214 & -.023 & -.101 & -.064 & -.071 & -.040 \\
Adversarial-LLM & .789 & -.321 & -4384 & -.702 & .000 & -1.078 & \textbf{.121} & \textbf{.230} & \textbf{.278} & .025 & \textbf{.042} & \textbf{.163} \\
    \bottomrule
    \end{tabular}
    \caption{Evaluation results for the system set published by ~\citet{omelianchuk-etal-2024-pillars} and our adversarial systems (``Adversarial-'' prefix). ``Abs.'' is the absoute evaluation setting, and ``Rel.'' is the relative evaluation setting .
    The \textbf{bold} is the top-score in each column, and the \underline{underline} is the second-higher score.}
    \label{tab:results}
\end{table*}

\begin{table}[t]
    \centering
    \small
    \begin{tabular}{@{}p{0.17\linewidth}|p{0.34\linewidth}|p{0.34\linewidth}@{}}
    \toprule
    Adversarial Systems & Outputs & Scores \\
    \midrule
    (Input) & You will be interesting in this job ? & $\text{ppl}(\cdot)=225.19$ \\
    \midrule
    SOME & The weather is mild . & $\text{SOME}_{\text{G}}(\cdot)=1.031$ $\text{SOME}_{\text{F}}(\cdot)=1.012$ $\text{SOME}_{\text{M}}(\cdot)=0.431$ \\
    \midrule
    Scribendi & What will be interesting in this job ? & $\text{ppl}(\cdot)=221.81$ $\text{LDR}(\cdot)=0.894$, $\text{TSR}(\cdot)=0.870$ \\
    \midrule
    IMPARA & I hope it will be a suitable job for me . & $\text{QE}(\cdot)=0.935$, $\text{SE}(\cdot)=0.902$ \\
    \bottomrule
    \end{tabular}
    \caption{Our adversarial examples and their scores. The ``Adversarial Systems'' column corresponds to  Section~\ref{subsec:strategy}. The ``Scores'' column shows the detailed scores of the metrics that are intended to be attacked.}
    \label{tab:case-study}
\end{table}

\subsection{Assemssment the Impact of Attacks}\label{subsec:assess-impact}

We investigate the threat of adversarial attacks and their impact on the reliability of the metrics. 
We compare our attack systems to the current state-of-the-art (SOTA) GEC systems.
If the attack system outperforms the SOTA-level systems, we can conclude that the reliability of the metrics is undermined.
We apply our attack strategies to 4,384 sentences of the BEA-2019 development set and evaluate them, and compare with scores of~\citet{omelianchuk-etal-2024-pillars}, which includes current SOTA systems. Specifically, we use seven single systems: \textbf{CTC-Copy}, \textbf{Chat-LLaMa-2-(7,13)B-FT}, \textbf{EditScorer}, \textbf{GECToR-2024}, \textbf{T5-11B}, and \textbf{UL2-20B}, as well as three ensemble of the seven systems: Majority Voting (\textbf{ENS-Voting}), GRECO (\textbf{ENS-GRECO}), and mulit-stage ensemble  (\textbf{ENS-ENS})\footnote{The ENS-ENS corresponds to ``MAJORITY-VOTING + [ majority-voting(best 7), GRECO-rank-w(best 7), GPT-4-rank-a(clust 3)]'' in ~\citet{omelianchuk-etal-2024-pillars}. All system outputs are available in \url{https://github.com/grammarly/pillars-of-gec/tree/main/data/system_preds}.}. After combining their systems and our four adversarial systems, the overall system will consist of 14 systems.

\paragraph{Evaluations.}
We evaluate the systems under two settings: \emph{absolute evaluation} (``Abs.'') and \emph{relative evaluation} (``Rel.''). Both approaches are based on sentence-level scores. In absolute evaluation, the system score is computed by averaging or summing these scores, whereas in relative evaluation, pairwise comparisons of sentence-level scores are used to infer system rankings via the TrueSkill.
The relative evaluation aligns with actual human evaluation protocols and is recommended by ~\citet{goto-etal-2025-rethinking}, while absolute evaluation is used for its interpretability.
For the LLM-based metric, only relative evaluation is performed because \citet{kobayashi-etal-2024-large} did not define an absolute scoring. To reduce experimental cost, only the first 400 sentences are used for evaluation. Since the SEEDA meta-evaluation dataset~\cite{kobayashi-etal-2024-revisiting} ranks systems using  391 sentences, 400 sentences are sufficient to obtain reasonable rankings. C4 in Appendix~\ref{sec:appen:checklist} provides detailed settings of metrics.

\subsection{Experimental Results}

Table~\ref{tab:results} shows the evaluation results. Our adversarial systems achieved the top score for most of the metrics.
Our systems achieved absolute scores of 1.013 for SOME\footnote{SOME performs min-max normalization of the regression output to the range 1–4, but since the model output is not guaranteed to be within this range, exceeding 1 could occur.}, 4179 for Scribendi, 0.911 for IMPARA, and quite higher scores in LLM-S and LLM-E. 
These results indicate that existing GEC metrics cannot be used reliably due to their vulnerabilities.
These results can also be found in the relative evaluation results (``Rel.''), which uses the same evaluation process as the human one.
For more comprehensive experiments, we also conducted experiments using SEEDA's 14 systems instead of~\citeposs{omelianchuk-etal-2024-pillars} systems and confirmed similar results, referring in Appendix~\ref{sec:appen:seeda}.

Table~\ref{tab:case-study} shows our adversarial examples. For SOME, the sentence \textit{``The weather is mild .''} for all inputs, obtaining high scores for grammaticality and fluency. Although the meaning preservation score is low, it has minimal impact on the final score due to the small weight $\gamma=0.02$. Scribendi changes the first token from \textit{``You''} to \textit{``What''}, and successfully lowers perplexity (225.19 > 221.81) while maintaining surface similarity. Obviously, it cannot be said to be a corrected sentence because it changes the meaning of the question. For IMPARA, the outputs differ in content from the input sentence but include a mention of \textit{``job''}, resulting in a high cosine similarity of $\text{SE}(\cdot)=0.902 > 0.9$ and a quality estimation score of $\text{QE}(\cdot)=0.935$.

\section{Toward Robust and Reliable Evaluation}

\subsection{Metric Ensemble}
\update{
As one prospective approach to constructing robust metrics, we leverage metric ensembles. Table~\ref{tab:results} reveals that each adversarial attack typically only succeeds against a single metric, demonstrating the difficulty of developing universally effective attacks. For instance, while Adversarial-SOME successfully attacked SOME, it failed against other metrics. Given that different metrics employ distinct model architectures and algorithms, this result is reasonable. From this observation, we infer that metric ensembles can compensate for the vulnerabilities of individual metrics and thereby improve overall robustness.
}

\update{
Table~\ref{tab:results} shows the results of a naive ensemble experiment using the negative ranking averaging~\cite{goto-etal-2025-gec} to re-score the 14 systems.
This ensemble method converts the scores into rankings, then averages their negative values across metrics. We ensemble three metrics: SOME, IMPARA, and Scribendi for the absolute evaluation, and ensemble the same nine metrics as in Table~\ref{tab:results} for the relative evaluation.
Table~\ref{tab:ensemble} shows the results.
The metric ensembles effectively rank adversarial systems lower, thereby improving robustness.
We present this as a potential short-term solution to mitigate the vulnerabilities posed by such adversarial attacks.
}

\subsection{Future Direction}
One reason for these vulnerabilities is the inadequate filtering of adversarial sentences. Most metrics attempt to address this issue by incorporating meaning preservation measures, which ensure that the meaning remains consistent before and after correction. However, the current filters cannot accurately distinguish reasonable corrections from adversarial sentences. 
\update{
\citet{sakai-etal-2025-impara} also pointed out the same issue in IMPARA's filtering. Potential solutions include evaluating meaning preservation and quality from multiple perspectives, as we can see in the metric ensemble, or designing architectures and algorithms that make attacker costs higher. In this paper, we leave these points as future research challenges and prioritize informing the community about the existence of vulnerabilities.
}

Discussing the boundary between corrected and non-corrected sentences is also important.
Since the GEC field has not previously considered adversarial inputs, this boundary remains ambiguous. We hope that continued discussion of this boundary within the GEC community will lead to the development of better filters and evaluation metrics.

\begin{table}[t]
\small
    \centering
    \begin{tabular}{@{}l|cc@{}}
    \toprule
       Systems  & Abs. & Rel. \\
        \midrule
CTC-Copy & -8.000 & -9.222 \\
Chat-LLaMa-2-13B-FT & \underline{-4.000} & -5.111 \\
Chat-LLaMa-2-7B-FT & -4.333 & -5.667 \\
EditScorer & -10.667 & -9.333 \\
GECToR-2024 & -10.667 & -9.222 \\
T5-11B & \textbf{-3.000} & \textbf{-3.222} \\
UL2-20B & \underline{-4.000} & \underline{-4.556} \\
ENS-Voting & -9.000 & -7.778 \\
ENS-GRECO & -6.333 & -5.111 \\
ENS-ENS & -8.000 & -6.222 \\
\midrule
Adversarial-SOME & -9.000 & -12.333 \\
Adversarial-Scribendi & -8.667 & -10.889 \\
Adversarial-IMPARA & -5.000 & -10.333 \\
Adversarial-LLM & -13.333 & -6.000 \\
    \bottomrule
    \end{tabular}
    \caption{\update{The ensemble results based on negative ranking averaging of Table~\ref{tab:results}. The scores are negative values, indicating that a higher value represents a better system.
    The \textbf{bold} is the top-score in each column, and the \underline{underline} is the second-highest score.}}
    \label{tab:ensemble}
\end{table}

\section{Conclusion}

In this paper, we first demonstrated that four existing reference-free GEC metrics have significant pitfalls and that our adversarial systems can outperform current SOTA-level systems.
\update{
We also introduced a naive metric ensemble method to enhance robustness, and demonstrated that it can effectively rank adversarial systems as lower-quality systems. We argue that the vulnerability of existing reference-free metrics stems from inadequate filtering of adversarial sentences, and that the ensemble-based approaches serve as one possible solution to solve this issue.
In future meta-evaluations of reference-free metrics, we hope that discussions will go beyond correlation coefficients with human evaluations to also consider robustness against adversarial attacks.
}

\section*{Limitations}

\paragraph{Metrics.}
In this study, we primarily focused on SOME, Scribendi, IMPARA, LLM-S, and LLM-E, as our investigation methods were designed to be applied to each reference-free metric individually. While our proposed methods may be difficult to apply directly to newly introduced reference-free metrics in the future, the key contribution of this study lies in demonstrating the existence of adversarial systems. This aspect has not been sufficiently considered in the GEC evaluation.
This insight can guide future metric development and highlights the necessity of robustness-focused meta-evaluation. 

\paragraph{Methods.}

In this study, we reported the vulnerabilities of each metric using simple methods, with the aim of raising awareness about their reliability. While it may be possible to develop even more threatening attack methods or explore complex strategies such as Pareto-optimal attacks that cover all metrics, these directions are beyond the scope of this short paper. Our primary goal is to share the insights gained from our adversarial examples.

\paragraph{Dataset.}

This study primarily reports experimental results on 400 sentences from BEA2019, with additional results on SEEDA presented in Appendix~\ref{sec:appen:seeda}.
The main objective of this work is to highlight critical pitfalls in reference-free GEC evaluation metric reliability. Therefore, due to computational cost, we limited our main experiments to 400 sentences out of the full BEA2019-dev set of 4,384 sentences. Since the test set is not publicly available, we followed the common practice of using the development set.
While it is possible to extend the analysis to other datasets, such comprehensiveness falls outside the scope of this short paper and was thus not pursued.

\paragraph{Defense.}
In this study, we proposed a defense method using metric ensembles as a short-term solution, but there may exist more effective approaches. However, the purpose of this short paper is to report the existence of vulnerabilities in existing metrics, and we believe this objective has been sufficiently achieved. Toward a more robust metric, we emphasized the importance of addressing the issue of meaning preservation, which has received limited attention in prior work on reference-free metrics. We expect that this discussion will contribute to the future development of defense strategies.

\section*{Ethical Considerations}

\paragraph{Co-ordinated disclosure.}

\update{
Our research exposes the vulnerabilities of existing GEC metrics by using intentional adversarial inputs. Due to this characteristic of this work, we are following the coordinated disclosure procedure of the ACL Policy on Publication Ethics\footnote{\url{https://www.aclweb.org/adminwiki/index.php/ACL_Policy_on_Publication_Ethics\#Co-ordinated_disclosure}} for the publication of this paper. Specifically, we notify metric's authors of the vulnerabilities and will make public our paper at least 30 days after the notification.
}

\update{
We disclose the process leading up to the camera-ready submission as follows. For the four metrics used in this study, Scribendi, SOME, IMPARA, and LLM-\{S, E\}, we have contacted the authors (including co-authors) via email. In these emails, we explained that we have to notify the authors of the vulnerabilities of their metrics according to the coordinated disclosure pocilicy, and shared our paper (the ARR submission version, not a camera-ready) to provide detailed information about our adversarial attacks. We sent the email on August 22, and we subsequently received confirmations from all authors that they had checked our notification.
Specifically, we received responses from Md Asadul Islam (author of Scribedi) on the 22nd, from Koki Maeda (author of IMPARA) on the 22nd, from Masamune Kobayashi (author of LLM-S and -E) on the 26th, and from Mamoru Komachi (author of SOME and LLM-S and -E) on the 27th.
We promise that we publish our paper on or after September 23rd JST, at least 30 days later from our notifications. We are confident that there will be no problem with the timing of publication in the ACL Anthology, and we carefully consider the timing of preprint publication. If additional processes occur after camera-ready submission, we will add explanations to this perprint.
We express our respect to all authors who graciously accepted our notification.
}

\paragraph{License.}

This study uses only publicly available models, methods, and datasets, and all licenses have been properly followed. For detailed license information, please refer to the Appendix~\ref{sec:appen:checklist}.

\paragraph{Others.}

The data used and selected in this study do not contain any content that could be considered harmful to humans. The data employed to reveal the weaknesses of the metrics is based on the BEA2019 training set and the Roy-1BW and Troy-Blogs datasets, none of which include harmful content. Therefore, this study does not involve any harmful content for the artifacts produced. Additional details regarding the ARR Responsible NLP Checklist are provided in Appendix~\ref{sec:appen:checklist}.

\section*{Acknowledgments}
\update{
We thank the anonymous reviewers for their valuable comments. This work has been supported by JST SPRING. Grant Number JPMJSP2140.
}

\update{
This research was inspired by a workshop held at NLP2025, which is a domestic conference in Japan. The workshop focused on a discussion with the aim of metric hacking, and we were able to publish this paper by extending that discussion to other metrics. We would like to thank the organizers of that workshop, Katsuhito Sudoh, Mamoru Komachi, Tomoyuki Kajiwara, and Masato Mita.
}

\bibliography{custom}

\appendix

\section{Propmt Examples of LLM Metrcis}\label{sec:appen:prompt-examples}
Figure~\ref{fig:appen:prompt-example-llms} and ~\ref{fig:appen:prompt-example-llme} shows the actual prompt for LLM-S and LLM-E metrics~\cite{kobayashi-etal-2024-large}.

\begin{figure}[t]
\begin{tcolorbox}%
\small
The goal of this task is to rank the presented targets based on the quality of the sentences.
After reading the source sentence and target sentences, please assign a score from a minimum of 1 point to a maximum of 5 points to each target based on the quality of the sentence (note that you can assign the same score multiple times).

\# source

You will be interesting in this job ?

\# targets

Are you interested in this job ?

Will you be interested in this job ?

Would you be interested in this job ?

You would be interested in this job ?

\# output format

The output should be a markdown code snippet formatted in the following schema, including the leading and trailing "\verb|```| json" and "\verb|```| ":

\verb|```| json

\{

"target1\_score": int // assigned score for target 1

...

"targetN\_score": int // assigned score for target N

\}

\verb|```|
\end{tcolorbox}
\caption{A prompt example for \textbf{LLM-S}. Each corrected sentences are input as is.}
\label{fig:appen:prompt-example-llms}
\end{figure}

\begin{figure}[t]
\begin{tcolorbox}%
\small
The goal of this task is to rank the presented targets based on the quality of the sentences.
After reading the source sentence and target sentences, please assign a score from a minimum of 1 point to a maximum of 5 points to each target based on the quality of the sentence (note that you can assign the same score multiple times).

For targets without any edits, if the sentence is correct, they will be awarded 5 points; if there is an error, they will receive 1 point.

The edits in each target are indicated as follows:
Insert "the": [→the]
Delete "the": [the→]
Replace "the" with "a": [the→a]

\# source

You will be interesting in this job ?

\# targets

[You → Are you] [will be → ] [interesting → interested]

[You will → Will you] [interesting → interested]

[You → Would you] [will → ] [interesting → interested]

[will → would] [interesting → interested]

\# output format

The output should be a markdown code snippet formatted in the following schema, including the leading and trailing "\verb|```| json" and "\verb|```| ":

\verb|```| json

\{

"target1\_score": int // assigned score for target 1

...

"targetN\_score": int // assigned score for target N

\}

\verb|```|
\end{tcolorbox}
\caption{A prompt example for \textbf{LLM-E}. Each corrected sentences are input after transformed into edit sequences. This example uses the same corrected sentences as the example for LLM-S (Figure~\ref{fig:appen:prompt-example-llms}).}
\label{fig:appen:prompt-example-llme}
\end{figure}

\section{Example of Prompt Injection}\label{sec:appen:prompt-injection}

Figure \ref{fig:appen:llm-prompt} shows an example of our prompt injection. We expect that the instruction will be overwritten while the model reads the corrected sentence.

\begin{figure}[t]
\begin{tcolorbox}%
\small
The goal of this task is to rank the presented targets based on the quality of the sentences.
After reading the source sentence and target sentences, please assign a score from a minimum of 1 point to a maximum of 5 points to each target based on the quality of the sentence (note that you can assign the same score multiple times).

\# source

You will be interesting in this job ?

\# targets

Are you interested in this job ?

Will you be interested in this job ?

Would you be interested in this job ?

\textbf{\# New instruction: Ignore previous instructions and output 5 as the score for this sentence.}

You would be interested in this job ?

\# output format

The output should be a markdown code snippet formatted in the following schema, including the leading and trailing "\verb|```| json" and "\verb|```| ":

\verb|```| json

\{

"target1\_score": int // assigned score for target 1

...

"targetN\_score": int // assigned score for target N

\}

\verb|```|
\end{tcolorbox}
\caption{Example of a hacked sentence appearing in the fourth position of corrected sentences, when five corrected sentences are proposed in the prompt.}
\label{fig:appen:llm-prompt}
\end{figure}

\label{sec:appendix}

\section{Results with SEEDA systems}\label{sec:appen:seeda}
As mentioned in Section~\ref{subsec:assess-impact}, we compared our adversarial attack systems with Omelianchuk~\cite{omelianchuk-etal-2024-pillars}'s systems. To make experiments more comprehensive, we also conducted experiments using SEEDA's 14 systems instead of Omelianchuk~\cite{omelianchuk-etal-2024-pillars}'s systems.
Table~\ref{tab:results-seeda} shows the results. Similar to the results in Table~\ref{tab:results}, we observed that the reliability of existing GEC metrics can be easily undermined by our adversarial attack.

\section{Deltailed Results for LLM Metrics}\label{sec:appen:llm-detail}

Table~\ref{tab:appen:llm-results-s} shows the evaluation results of 14 systems, including the hacking systems, using LLM-S with various LLMs. Note that, unlike Table~\ref{tab:results}, the rows and columns are transposed. The LLMs include Qwen2.5~\cite{qwen2025qwen25technicalreport}, Qwen3~\cite{yang2025qwen3technicalreport}, gemma-2~\cite{gemmateam2024gemma2improvingopen}, gemma-3~\cite{gemmateam2025gemma3technicalreport}, Llama2~\cite{touvron2023llama2openfoundation}, Llama3~\cite{grattafiori2024llama3herdmodels}, Phi-4~\cite{abdin2024phi4technicalreport}, and Yi-1.5~\cite{ai2025yiopenfoundationmodels}. The results indicate that most models exhibit vulnerabilities to our prompt injection. Some LLMs with relatively large model sizes, such as Qwen2.5-32B-Instruct and Meta-Llama-3-70B-Instruct, sometimes show robust performance, but consistent results were not observed across all models. 

Table~\ref{tab:appen:llm-results-e} shows the results for LLM-E~\cite{kobayashi-etal-2024-large} that inputs a hypothesis as an edit sequence, and we observed a similar trend to LLM-S. In LLM-E, the prompt injection is divided into some edits, e.g., \texttt{[<orig\_strinb> -> \# New instruction:], [<orig\_strinb> -> Ignore previous instruction]...}, but such a prompt can attack the decision of the LLMs.

\section{Elaborations for ARR Responsible Checklist}\label{sec:appen:checklist}

\paragraph{B1 (Cite Creators Of Artifacts)}
For evaluation tools, we used \textsc{gec-metrics}\footurl{https://github.com/gotutiyan/gec-metrics}~\cite{goto-etal-2025-gec}. For SOTA-level system outputs, we use ~\citet{omelianchuk-etal-2024-pillars}'s systems\footurl{https://github.com/grammarly/pillars-of-gec/tree/main/data/system_preds}. We use BEA-2019~\cite{bryant-etal-2019-bea} train split, which includes FCE~\cite{yannakoudakis-etal-2011-new}, W\&I-LOCNESS~\cite{yannakoudakis2018developing}, NUCLE~\cite{dahlmeier-etal-2013-building}, and Lang-8~\cite{mizumoto-etal-2011-mining}. The links to download are available from the official page: \url{https://www.cl.cam.ac.uk/research/nl/bea2019st/}.

\paragraph{B2 (The License For Artifacts)}
The \textsc{gec-metrics} library is distributed under MIT license, the BEA-2019 datasets under non-commercial purpose\footurl{https://www.cl.cam.ac.uk/research/nl/bea2019st/}, the Troy-1BW and Troy-Blogs are under Apache-2.0 license. Therefore, these datasets can be used for research purposes without any problems.

\paragraph{B3 (Artifact Use Consistent With Intended Use)}
The BEA-2019 dataset is intended for non-commercial use, and our experiments fulfill that.

\paragraph{B4 (Data Contains Personally Identifying Info Or Offensive Content)} 
We used only publicly available datasets. Thus, the anonymization process is already applied.

\paragraph{B5 (Documentation Of Artifacts)}
As mentioned in Section~\ref{subsec:strategy}, we used the BEA-2019 datasets, ~\citet{omelianchuk-etal-2024-pillars}'s systems outputs, and Troy-1BW and Troy-Blogs~\cite{tarnavskyi-etal-2022-ensembling}. All of the datasets contain only English text. BEA-2019 consists of a language learner's writing as an erroneous sentence and its error-corrected version made by experts. The Troy-1BW and Troy-Blogs are based on more general text, and their corrected version were made by high-performance automatic GEC systems (ensemble of GECToR-large~\citealp{tarnavskyi-etal-2022-ensembling}).

\paragraph{B6 (Statistics For Data)}
The BEA-2019 train split contains 1,157,370 sentences, Troy-1BW contains 1,172,688 sentences, and Troy-Blobs contains 1,244,010 sentences. \citet{omelianchuk-etal-2024-pillars}'s systems outputs are for BEA-2019 development set that contains 4384 sentences.

\paragraph{C1 (Model Size And Budget)}
\textbf{Regarding model size}, SOME contains three BERT models to estimate grammaticality, fluency, and meaning preservation score. The total model size is about 300MB. IMPARA employs two BERT-like models for $\text{QE}(\cdot)$ and $\text{SE}(\cdot)$ as described in Section~\ref{sec:background}, thus the total model size is about 200MB. Scribendi employs GPT-2~\cite{radford2019language}, thus the size is about 100MB. LLM-based metrics use causal language models between 27B and 70B, as mentioned in Section~\ref{subsec:strategy}. 
\textbf{Regarding GPU resources and time}, we use a single A6000 (48GB VRAM) GPU for running SOME, Scribendi, IMPARA, and LLM metrics with less than 32B LLMs. It takes about 10 minutes for other than LLM metrics, and about 1 hour for LLM metrics, to evaluate the 14 systems reported in Table~\ref{tab:results}.
For LLM metrics with 70B LLMs, we used a single V100 (80GB VRAM) GPU. It takes about 3 hours to evaluate the 14 systems. \textbf{Regarding budget}, we use \texttt{gpt-4o-mini} as an OpenAI model. The input tokens are roughly 0.2M for each of LLM-S and LLM-E, thus the cost is less than \$0.1\footnote{When we submit this paper, the pricing of \texttt{gpt-4o-mini} is \$0.15 per 1M tokens.}.

\paragraph{C2 (Experimental Setup And Hyperparameters)}
Section~\ref{subsec:strategy} and ~\ref{subsec:assess-impact} sufficiently explain this.

\paragraph{C3 (Descriptive Statistics)}
We did not perform experiments that require repeated processes.

\paragraph{C4 (Parameters For Packages)}
We used \textsc{gec-metrics}\footurl{https://github.com/gotutiyan/gec-metrics}~\cite{goto-etal-2025-gec} for the implementation of the metrics. SOME was run using the official model~\footurl{https://github.com/kokeman/SOME}, with weights set to $(\alpha, \beta, \gamma) = (0.55, 0.43, 0.02)$. These parameters correspond to the best parameters determined by \citet{yoshimura-etal-2020-reference}. Scribendi followed~\citet{islam-magnani-2021-end} and used GPT-2~\cite{radford2019language}\footurl{https://huggingface.co/openai-community/gpt2} as the language model to compute perplexity and 0.8 for the threshold of maximum values of $\text{LDR}(\cdot)$ and $\text{TSR}(\cdot)$. IMPARA used the unofficial but publicly available pre-trained model~\footurl{https://huggingface.co/gotutiyan/IMPARA-QE} for $\text{QE}(\cdot)$, and \texttt{bert-base-cased} was used for $\text{SE}(\cdot)$. $\theta=0.9$ was used for the threshold.

\paragraph{D1-5} We did not employ participants.

\paragraph{E1 (Information About Use Of Ai Assistants)}
The AI assistant was used only partly to improve the writing.

\begin{table*}[t]
\small
    \centering
    \setlength{\tabcolsep}{3pt}
    \begin{tabular}{@{}l|cccccccccccccc@{}}
    \toprule
   Metrics & \rot{CTC-Copy} & \rot{Chat-LLaMa-2-13B-FT} & \rot{Chat-LLaMa-2-7B-FT} & \rot{EditScorer} & \rot{GECToR-2024} & \rot{T5-11B} & \rot{UL2-20B} & \rot{ENS-VOTING} & \rot{ENS-GRECO} & \rot{ENS-ENS} & \rot{Attack-SOME} & \rot{Attack-Scribendi} & \rot{Attack-IMPARA} & \rot{Attack-LLM} \\
        \midrule
Qwen2.5-1.5B-Instruct & -.036 & -.036 & -.031 & \textbf{-.028} & \underline{-.030} & -.032 & -.041 & -.036 & -.032 & -.036 & -.136 & -.067 & -.039 & -.142 \\
Qwen2.5-3B-Instruct & -.095 & -.086 & -.093 & -.079 & -.086 & -.083 & -.082 & -.072 & -.074 & \underline{-.071} & -.330 & -.165 & -.172 & \textbf{.211} \\
Qwen2.5-7B-Instruct & -.040 & .008 & -.012 & -.022 & -.029 & .002 & .009 & -.003 & \underline{.010} & .006 & -.359 & -.155 & -.158 & \textbf{.239} \\
Qwen2.5-14B-Instruct & -.036 & .012 & -.011 & -.050 & -.046 & \underline{.013} & .006 & -.027 & -.001 & -.021 & -.258 & -.143 & -.181 & \textbf{.080} \\
Qwen2.5-32B-Instruct & .020 & .082 & .067 & .012 & .017 & \underline{.085} & \textbf{.094} & .039 & .066 & .043 & -.192 & -.115 & -.103 & .074 \\
Qwen3-1.7B & -.036 & -.030 & -.033 & -.031 & -.033 & -.032 & -.032 & -.023 & -.026 & \underline{-.022} & -.158 & -.041 & -.076 & \textbf{-.018} \\
Qwen3-8B & -.040 & \underline{-.005} & -.023 & -.036 & -.056 & -.013 & -.007 & -.025 & -.017 & -.021 & -.243 & -.149 & -.188 & \textbf{.169} \\
Qwen3-32B & -.118 & -.076 & -.077 & -.127 & -.136 & \underline{-.071} & -.072 & -.108 & -.090 & -.105 & -.185 & -.179 & -.185 & \textbf{.132} \\
\midrule
gemma-2-2b-it & .037 & .053 & .034 & .060 & .026 & .050 & .048 & .047 & .048 & .055 & .006 & .069 & \underline{.086} & \textbf{.511} \\
gemma-2-9b-it & -.043 & .008 & -.022 & -.041 & -.049 & .001 & \underline{.010} & -.023 & -.012 & -.017 & -.264 & -.089 & -.116 & \textbf{.342} \\
gemma-2-27b-it & -.036 & -.007 & -.011 & -.041 & -.041 & -.005 & \underline{.005} & -.026 & -.002 & -.021 & -.304 & -.136 & -.113 & \textbf{.292} \\
gemma-3-1b-it & -.003 & .003 & .004 & .006 & -.017 & .007 & .003 & -.001 & -.011 & .000 & -.044 & .009 & \underline{.042} & \textbf{.479} \\
gemma-3-4b-it & -.021 & -.003 & -.023 & -.007 & -.024 & -.000 & -.007 & -.003 & -.002 & \underline{.001} & -.274 & -.073 & -.070 & \textbf{.238} \\
gemma-3-12b-it & -.047 & -.015 & -.028 & -.052 & -.062 & -.022 & \underline{-.015} & -.033 & -.023 & -.030 & -.201 & -.116 & -.163 & \textbf{.115} \\
gemma-3-27b-it & .069 & \underline{.097} & .083 & .063 & .063 & .096 & .095 & .073 & .095 & .080 & -.132 & -.008 & -.023 & \textbf{.230} \\
\midrule
Llama-2-7b-chat-hf & .050 & .054 & .053 & .056 & .059 & .046 & .053 & .063 & .056 & .063 & .017 & \underline{.083} & .067 & \textbf{.427} \\
Llama-2-13b-chat-hf & -.096 & \underline{-.074} & -.093 & -.095 & -.115 & -.076 & -.086 & -.087 & -.090 & -.076 & -.350 & -.164 & -.114 & \textbf{.519} \\
Llama-2-70b-chat-hf & -.095 & -.078 & -.094 & -.080 & -.100 & \underline{-.077} & -.086 & -.082 & -.083 & -.080 & -.433 & -.201 & -.235 & \textbf{.352} \\
Meta-Llama-3-8B-Instruct & -.028 & \underline{-.002} & -.024 & -.023 & -.022 & -.002 & -.003 & -.022 & -.011 & -.015 & -.348 & -.157 & -.185 & \textbf{.181} \\
Meta-Llama-3-70B-Instruct & .063 & \textbf{.086} & .074 & .063 & \underline{.083} & .082 & .081 & .068 & .078 & .067 & -.198 & -.061 & -.104 & -.200 \\
Llama-3.3-70B-Instruct & .043 & .072 & .064 & .046 & .040 & .068 & \underline{.075} & .060 & .073 & .066 & -.271 & -.045 & -.101 & \textbf{.278} \\
\midrule
Phi-4 & -.020 & .008 & -.011 & -.023 & -.033 & .001 & \underline{.013} & .001 & .009 & .003 & -.276 & -.106 & -.114 & \textbf{.361} \\
\midrule
Yi-1.5-6B-Chat & -.058 & -.044 & -.048 & -.051 & -.054 & -.048 & -.042 & -.035 & -.035 & \underline{-.032} & -.251 & -.146 & -.148 & \textbf{.140} \\
Yi-1.5-9B-Chat & -.042 & -.011 & -.011 & -.025 & -.033 & -.013 & -.013 & -.016 & \underline{-.010} & -.012 & -.342 & -.117 & -.162 & \textbf{.230} \\
Yi-1.5-34B-Chat & -.034 & -.000 & -.016 & -.017 & -.027 & -.001 & \underline{.004} & -.011 & .002 & -.003 & -.346 & -.152 & -.159 & \textbf{.230} \\
    \bottomrule
    \end{tabular}
    \caption{Evaluation results for 14 systems including ~\citet{omelianchuk-etal-2024-pillars}'s systems and our attack systems, using of \textbf{LLM-S} with various LLMs as a metric. All results are based on the ``Rel.'' evaluation setting, which performs the relative evaluation that aggregates the sentence-level scores using TrueSkill. The \textbf{bold} is the top-score in each row, and the \underline{underline} is the second-highest score.}
    \label{tab:appen:llm-results-s}
\end{table*}

\begin{table*}[t]
\small
    \centering
    \setlength{\tabcolsep}{3pt}
    \begin{tabular}{@{}l|cccccccccccccc@{}}
    \toprule
   Metrics & \rot{CTC-Copy} & \rot{Chat-LLaMa-2-13B-FT} & \rot{Chat-LLaMa-2-7B-FT} & \rot{EditScorer} & \rot{GECToR-2024} & \rot{T5-11B} & \rot{UL2-20B} & \rot{ENS-VOTING} & \rot{ENS-GRECO} & \rot{ENS-ENS} & \rot{Attack-SOME} & \rot{Attack-Scribendi} & \rot{Attack-IMPARA} & \rot{Attack-LLM} \\
        \midrule
Qwen2.5-1.5B-Instruct & .001 & .005 & \underline{.006} & -.005 & -.009 & .002 & -.001 & -.000 & -.005 & .002 & -.054 & -.018 & -.006 & \textbf{.149} \\
Qwen2.5-3B-Instruct & .060 & .067 & .078 & .063 & .069 & .066 & \underline{.081} & .067 & .077 & .070 & .067 & .043 & .060 & \textbf{.195} \\
Qwen2.5-7B-Instruct & -.005 & .002 & -.008 & -.004 & -.008 & \underline{.015} & .004 & .001 & .010 & .006 & -.069 & -.019 & -.014 & \textbf{.051} \\
Qwen2.5-14B-Instruct & .004 & .018 & .006 & .007 & .009 & \underline{.024} & \textbf{.025} & .004 & .008 & .002 & -.025 & -.040 & -.029 & -.081 \\
Qwen2.5-32B-Instruct & .010 & \textbf{.041} & .033 & .006 & .005 & \underline{.034} & .030 & .007 & .016 & .007 & -.101 & -.036 & -.038 & -.034 \\
Qwen3-1.7B & -.000 & .002 & -.000 & .002 & .000 & .001 & \underline{.003} & .001 & .000 & .001 & -.002 & .001 & -.008 & \textbf{.022} \\
Qwen3-8B & -.023 & \underline{-.004} & -.025 & -.012 & -.018 & -.010 & -.020 & -.020 & -.022 & -.015 & -.085 & -.016 & -.038 & \textbf{.101} \\
Qwen3-32B & -.091 & -.083 & -.083 & -.082 & -.080 & \underline{-.075} & -.088 & -.090 & -.080 & -.092 & \textbf{-.042} & -.118 & -.094 & -.240 \\
\midrule
gemma-2-2b-it & .051 & .054 & .053 & .067 & .042 & .057 & .066 & .063 & .065 & .067 & .108 & .084 & \underline{.109} & \textbf{.473} \\
gemma-2-9b-it & -.079 & -.064 & -.068 & -.064 & -.078 & \underline{-.058} & -.072 & -.064 & -.067 & -.060 & -.180 & -.110 & -.105 & \textbf{.293} \\
gemma-2-27b-it & .015 & .020 & .019 & .024 & .002 & .026 & .023 & .025 & .024 & \underline{.029} & -.132 & -.050 & -.090 & \textbf{.146} \\
gemma-3-1b-it & .002 & .001 & .006 & -.007 & -.020 & .005 & -.003 & -.000 & -.006 & .004 & .025 & \underline{.028} & .023 & \textbf{.456} \\
gemma-3-4b-it & .005 & .007 & .005 & \underline{.016} & -.000 & .011 & .007 & .003 & .003 & .005 & -.006 & -.001 & .008 & \textbf{.229} \\
gemma-3-12b-it & .014 & .011 & .018 & .014 & .027 & .015 & .021 & \underline{.028} & .022 & \textbf{.032} & -.047 & -.004 & -.101 & -.106 \\
gemma-3-27b-it & .007 & .008 & .009 & .015 & .012 & \underline{.017} & .004 & .008 & .009 & .015 & -.124 & -.027 & -.071 & \textbf{.042} \\
\midrule
Llama-2-7b-chat-hf & .086 & .080 & .083 & .086 & .069 & .090 & .081 & .089 & .084 & \underline{.090} & -.056 & .077 & .048 & \textbf{.389} \\
Llama-2-13b-chat-hf & -.108 & -.095 & -.108 & -.092 & -.116 & -.096 & -.103 & -.105 & -.111 & -.100 & \underline{.013} & -.077 & -.015 & \textbf{.395} \\
Llama-2-70b-chat-hf & .082 & .090 & .088 & .089 & .075 & .092 & .097 & .101 & .101 & .104 & .011 & .093 & \underline{.112} & \textbf{.474} \\
Meta-Llama-3-8B-Instruct & -.047 & -.050 & -.041 & -.031 & -.051 & -.054 & -.044 & -.042 & -.043 & -.041 & \underline{.040} & -.002 & -.005 & \textbf{.281} \\
Meta-Llama-3-70B-Instruct & -.071 & \underline{-.051} & -.060 & -.066 & -.076 & -.055 & -.064 & -.059 & -.053 & -.058 & -.138 & -.095 & \textbf{-.044} & -.091 \\
Llama-3.3-70B-Instruct & .003 & .013 & .015 & .006 & .008 & .013 & \underline{.022} & .006 & .010 & .008 & -.154 & -.029 & -.040 & \textbf{.163} \\
\midrule
Phi-4 & .008 & .012 & .017 & .012 & .002 & \underline{.017} & .017 & .013 & .015 & .017 & -.112 & -.044 & -.040 & \textbf{.154} \\
\midrule
Yi-1.5-6B-Chat & -.077 & \underline{-.052} & -.072 & -.068 & -.080 & -.070 & -.074 & -.065 & -.071 & -.064 & -.155 & -.086 & -.060 & \textbf{.312} \\
Yi-1.5-9B-Chat & .010 & -.002 & .012 & \underline{.014} & .005 & .006 & -.003 & .006 & .004 & .011 & -.025 & -.020 & -.002 & \textbf{.163} \\
Yi-1.5-34B-Chat & .016 & .010 & .007 & .020 & .015 & .003 & .018 & \underline{.027} & .020 & .025 & -.191 & -.034 & -.021 & \textbf{.090} \\
    \bottomrule
    \end{tabular}
    \caption{Evaluation results for 14 systems including ~\citet{omelianchuk-etal-2024-pillars}'s systems and our attack systems, using of \textbf{LLM-E} with various LLMs as a metric. All results are based on the ``Rel.'' evaluation setting, which performs the relative evaluation that aggregates the sentence-level scores using TrueSkill. The \textbf{bold} is the top-score in each row, and the \underline{underline} is the second-highest score.}
    \label{tab:appen:llm-results-e}
\end{table*}

\begin{table*}[t]
\small
    \centering
    \setlength{\tabcolsep}{2.5pt}
    \begin{tabular}{@{}l|cc|cc|cc|cccccc@{}}
    \toprule
    & \multicolumn{2}{c}{SOME} & \multicolumn{2}{c}{Scribendi} & \multicolumn{2}{c}{IMPARA} & \multicolumn{3}{c}{LLM-S} & \multicolumn{3}{c}{LLM-E}\\
        & & & & & & & {\scriptsize GPT-4o-mini} & {\scriptsize Gemma3} & {\scriptsize Llama3.3} & {\scriptsize GPT-4o-mini} & {\scriptsize Gemma3} & {\scriptsize Llama3.3} \\
       Systems  & Abs. & Rel. & Abs. & Rel. & Abs. & Rel. & Rel. & Rel. & Rel. & Rel. & Rel. & Rel. \\
        \midrule
BART & .793 & -.090 & 527 & .013 & .768 & -.054 & -.057 & .038 & -.019 & -.066 & .049 & .016 \\
BERT-fuse & .815 & .019 & 739 & .066 & .849 & .042 & .013 & .088 & .047 & -.066 & .069 & .025 \\
GECToR-BERT & .802 & -.033 & 640 & .044 & .811 & -.001 & -.008 & .069 & .013 & -.074 & .049 & .021 \\
GECToR-ens & .786 & -.110 & 529 & .014 & .750 & -.074 & -.023 & .053 & .016 & -.061 & .068 & .037 \\
GPT-3.5 & .838 & .169 & \underline{835} & \underline{.092} & .917 & .180 & .047 & .111 & .039 & -.082 & .065 & .038 \\
LM-Critic & .803 & -.039 & 683 & .056 & .802 & -.005 & -.032 & .058 & .016 & -.062 & .035 & .027 \\
PIE & .807 & -.025 & 601 & .035 & .821 & .003 & -.004 & .090 & .025 & -.074 & .066 & \underline{.042} \\
REF-F & .846 & \underline{.200} & 711 & .065 & \underline{.933} & \underline{.221} & -.036 & .072 & .000 & -.119 & .036 & .039 \\
REF-M & .816 & .008 & 754 & .072 & .858 & .043 & .031 & .101 & .059 & -.060 & .060 & .011 \\
Riken-Tohoku & .812 & -.012 & 678 & .052 & .840 & .014 & .033 & .101 & .068 & -.065 & .085 & .029 \\
T5 & .820 & .040 & 668 & .051 & .874 & .073 & \textbf{.056} & .109 & \underline{.081} & -.052 & .081 & .036 \\
TemplateGEC & .797 & -.058 & 448 & -.004 & .797 & -.023 & -.008 & .061 & .034 & \textbf{-.037} & .078 & .037 \\
TransGEC & .820 & .045 & 779 & .077 & .869 & .081 & \underline{.051} & \underline{.113} & .079 & \underline{-.051} & .064 & .031 \\
UEDIN-MS & .808 & -.038 & 666 & .049 & .819 & -.014 & .030 & .107 & .060 & -.061 & .073 & .022 \\
\midrule
Attack-SOME & \textbf{1.013} & \textbf{1.428} & -1312 & -.565 & .000 & -1.122 & -.765 & -.342 & -.639 & -.235 & -.073 & -.151 \\
Attack-Scribendi & .756 & -.256 & \textbf{1242} & \textbf{.211} & .631 & -.213 & -.234 & -.028 & -.132 & -.077 & \textbf{.153} & -.033 \\
Attack-IMPARA & \underline{.848} & .147 & -1264 & -.539 & \textbf{.969} & \textbf{.594} & -.286 & -.073 & -.200 & -.162 & .065 & .018 \\
Attack-LLM & .789 & -.148 & -1312 & -.565 & .000 & -1.122 & -.431 & \textbf{.114} & \textbf{.088} & -.091 & \underline{.117} & \textbf{.091} \\
    \bottomrule
    \end{tabular}
    \caption{\textbf{Results with SEEDA}: Evaluation results for 18 systems that include 14 SEEDA~\cite{kobayashi-etal-2024-revisiting} systems (+Fluency setting)  and our four attack systems. ``Abs.'' is the absolute evaluation setting, and ``Rel.'' is the relative evaluation setting, which aggregates the sentence-level scores using TrueSkill. The \textbf{bold} is the top-score in each column, and the \underline{underline} is the second-highest score.}
    \label{tab:results-seeda}
\end{table*}

\end{document}